\documentclass{article}
\usepackage[utf8]{inputenc}
\usepackage{ascii}
\usepackage{newunicodechar}
\newunicodechar{♣}{\ENQ}
\newunicodechar{♠}{\ACK}
\usepackage{spconf,amsmath,graphicx}
\usepackage{booktabs} 
\usepackage{balance}
\usepackage{amsmath}
\usepackage{comment}
\usepackage{algorithmic}
\usepackage[ruled,vlined]{algorithm2e}
\usepackage{amssymb}
\usepackage{textcomp}
\usepackage{enumitem}
\usepackage{graphicx}
\usepackage{epstopdf}
\usepackage{wrapfig}
\usepackage{xspace}
\usepackage{subcaption}
\usepackage{array}
\usepackage{mathtools} 
\usepackage[hyphens]{url}
\usepackage{bm}
\usepackage{multirow}
\usepackage{adjustbox}
\usepackage{minitoc}
\usepackage[misc]{ifsym}
\usepackage{xcolor}

\title{SUVR: A Search-based Approach to Unsupervised\\Visual Representation Learning}
\name{♠ Yi-Zhan Xu, $\dagger$ Chih-Yao Chen, ♣ Cheng-Te Li}
\address{♠ ♣ Institute of Data Science, National Cheng Kung University\\
    $\dagger$ Institute of Information Science, Academia Sinica}

\begin{document}
%
\maketitle
\begin{abstract}
Unsupervised learning has grown in popularity because of the difficulty of collecting annotated data and the development of modern frameworks that allow us to learn from unlabeled data. Existing studies, however, either disregard variations at different levels of similarity or only consider negative samples from one batch. We argue that image pairs should have varying degrees of similarity, and the negative samples should be allowed to be drawn from the entire dataset. 
In this work, we propose Search-based Unsupervised Visual Representation Learning (SUVR) to learn better image representations in an unsupervised manner. 
We first construct a graph from the image dataset by the similarity between images, and adopt the concept of graph traversal to explore positive samples. In the meantime, we make sure that negative samples can be drawn from the full dataset. Quantitative experiments on five benchmark image classification datasets demonstrate that SUVR can significantly outperform strong competing methods on unsupervised embedding learning. Qualitative experiments also show that SUVR can produce better representations in which similar images are clustered closer together than unrelated images in the latent space.
\end{abstract}

\section{INTRODUCTION}
\label{sec:intro}
Deep learning has achieved astonishing performance due to the progress of computational power as well as the amount of data.
Although we could obtain satisfactory results with sufficient labeled data---from a few thousand to a few hundred thousand in general---human-annotated data is by no means economic.
Therefore, the idea of effectively leveraging unlabeled data, often refering to unsupervised or contrastive learning, has gained popularity recently~\cite{chen2020simple,Caron2018,wu2018unsupervised,Ye_2019,huang2019unsupervised}. 
Learning from data itself without the supervision of labels, however, remains a challenging task. Existing work for unsupervised contrastive learning on visual data takes the given image and its slightly varied version as positive pairs, while viewing all the other images as negatives~\cite{chen2020simple}. This assumes that all images other than the given image are equally negative. However, we argue that the distinct relations between images should also be considered, since the degree of similarities could vary a lot for different image pairs, e.g., \emph{tiger} v.s. \emph{leopard} apparently has a higher similarity than \emph{tiger} v.s. \emph{bird}.

In this work, we propose \textbf{\underline{S}earch-based \underline{U}nsupervised \underline{V}isual \underline{R}epresentation Learning} (SUVR), a novel framework that learns visual representations in an unsupervised manner.
First, we present the relations between images as a graph, and find each image's neighbors to learn better representations. Concretely, we utilize the concept of graph traversal to find positive samples--images that are similar in different aspects to a given image, as well as negative samples--images that are similar but should be separated. We then minimize the distances between positive pairs, and maximize the distances between negative pairs.
Moreover, we propose using three strategies to discover neighbors for each image, including (1)~Breadth-First Search (BFS), (2)~Depth-First Search (DFS) and (3)~Greedy Search, which reflect different aspects of information carried in the neighborhood. That is to say, BFS considers 1-hop information by finding the top-$k$ similar neighbors, while DFS considers up to $k$-hop information sequentially. 
Greedy search strikes the balance between BFS and DFS to incorporate the advantages of both. 
During training, we adopt the memory mechanism~\cite{Xiao_2017} to alleviate the over-smoothing problem~\cite{li2018deeper}, and it also allows us to draw negative samples from the entire dataset. 
We conduct experiments on five benchmark datasets for image classification task to evaluate the proposed SUVR. Experimental results show that SUVR consistently outperforms all the baselines by a large margin on every dataset. 
We also conduct qualitative studies to attest the quality of the learned representations. Results indicate that SUVR produces more visually meaningful embeddings compared with the most competitive baseline.
In short, our contribution is three-fold, as summarized below.
\begin{itemize}
    \item We propose a novel neighbor-discovering method to enhance unsupervised visual embedding learning, which effectively exploits graph traversal in an image graph.
    \item Three neighbor-discovering strategies, including BFS, DFS, and Greedy Search, are proposed to provide useful information that improves the robustness of image representations.
    \item Experiments conducted on five benchmark datasets show that SUVR gains substantial improvements, consistently surpassing the typical competing methods of unsupervised visual embedding learning. 
\end{itemize}

\section{METHODOLOGY}
\label{sec:model}

\subsection{Neighbor Discovering}
Given an unlabeled image set $\mathcal{D}$, we follow~\cite{wu2018unsupervised} to obtain the initial embedding for each instance. Next, a graph $\mathcal{G}=(\mathcal{V}, \mathcal{E})$ is constructed from $\mathcal{D}$, where each node $\mathcal{V}$ represents an image, and the edges $\mathcal{E}$ are the similarity between images.
After that, we propose three strategy for neighbor discovering.

\noindent\textbf{Breadth-First Search (BFS)}
is used to find top-$k$ neighbors. 
Specifically, we find $k$ most similar images for a given instance $x$: $\mathcal{N}^{B}_{pos}(x) = \{x\} \cup \{ x_{i} \; | \; x_{i} \ne x, \; s(x_{i}, x) \; \textrm{is} \; \textrm{top-}k \; \textrm{in} \; \mathcal{G} \}$, where $s$ is a function that calculates similarity between images following~\cite{wu2018unsupervised}.
In particular, $s$ measures the dot product similarity between the embedding of instance $x$ and the memory bank $\mathbf{M}$ of the whole image set $\mathcal{D}$. We select the top-$k$ images sorted by their similarities as the neighbors for $x$. BFS could be helpful because a variety of similar images are collected. For example, given an image of a tiger, we may find lynx, cheetah and puma as its neighbors. Here we denote neighbors found by BFS as $\mathcal{N}^{B}_{pos}$.

\noindent\textbf{Depth-First Search (DFS)} is used to find $k$-hop neighbors. For each iteration, we find the most similar image $j$ for $x$, and then we find the most similar image for $j$. This process is repeated for $k$ iteration. DFS aims to utilize information that could reach up to $k$-hop neighbors, exploring the image set in a depth-first manner. DFS could be useful because we can find different images that are somewhat similar. For example, given an image of warbler, we may sequentially find sparrow, pigeon and peacock. Here, peacock is found because it is the most similar image for pigeon. However, if we use BFS, we may not find peacock because it is less similar to warbler. The neighbor set found by DFS is denoted as $\mathcal{N}^{D}_{pos}(x)$.

\noindent\textbf{Greedy Search.}
Since BFS and DFS take distinct informational aspects into account, we provide a greedy strategy to merge them. The similarity gain specifically determines which strategy should be used in each iteration. We compare which strategy found the image with the highest degree of similarity for each iteration, and the strategy that produced the higher similar neighbor is chosen.
$x$: $\mathcal{N}^{G}_{pos}(x) = \{x\} \cup \{ x_{i} \; | \; x_{i} \ne x, \; max(BFS(\mathcal{G}), DFS(\mathcal{G})))$
\noindent\textbf{Negative Sampling.}
Different from~\cite{chen2020simple, Ye_2019} that consider negative samples only within a batch. We also aim to avoid the false separation problem~\cite{cao2016deep}. 
Thus, we adaptively choose the negatives from positive neighbors. 
Specifically, at each iteration, we select the least similar images from the positive set to form the negative samples $\mathcal{N}_{neg}$. In this way, the nagatives $\mathcal{N}_{neg}$ are expected to be more similar to the given image as the hard negatives~\cite{6909475,examplebased,felzenszwalb2010object}, and thereby help to learn more robust representations.

\subsection{Training Objective}
The objective function for SUVR is formulated as 
\begin{equation}
\begin{aligned}
\mathcal{L} = & -\sum_{i}\log{P(i|\mathbf{v})} \\
              & -\sum_{i} \sum_{j\in{\mathcal{N}_{pos}(x)}}\log{P(j|\mathbf{v})} \\
              & -\sum_{i} \sum_{k\in{\mathcal{N}_{neg}(x)}}\log{(1-P(k|\mathbf{v}))}
\label{eq:final}
\end{aligned}
\end{equation}
The first term is an instance-level loss that an image representation $\mathbf{v}$ should be assigned to the $i$-th instance, instead of the $i$-th class, calculated by
\begin{equation}
    P(i|\mathbf{v}) =  \frac{{\exp (\mathbf{M}^{\top}_{i} \mathbf{v} / \tau)}}{\sum_{j=1}^t \exp (\mathbf{M}^{\top}_{j} \mathbf{v} / \tau)} \label{eq:non_param}
\end{equation} 
Where $\mathbf{M}$ is the memory bank saving the updated embeddings, and $\tau$ is the temperature. 
The second term maximizes the likelihood between similar instances, bringing similar instances closer together. In the meantime, we separate the representations of unrelated instances in the final term to reduce the likelihood between negative pairs. Once we discover a new neighbor through the neighbor-discovery process, we calculate the loss and update the image representation.
\section{EXPERIMENTS}
\label{sec:experiments}
\subsection{Evaluation Settings}
\textbf{Datasets.}
We use five benchmark datasets with image classification task to evaluate the proposed SUVR for unsupervised embedding learning. The datasets are categorized as follows: \textbf{(1) Coarse-grained:} \emph{CIFAR-10}~\cite{article} consists of 10 classes. 50,000 images are available for training, and 10,000 photos are available for testing. \emph{SVHN}~\cite{37648} consists of 10 classes of digital house numbers. There are 73,257 samples for training, and 26,032 samples for testing. 
\textbf{(2) Fine-grained:} \emph{CIFAR-100} is a finer version of CIFAR-10 that sub-classes are also available. The sample size is the same as CIFAR-10 and there are 100 classes.
\emph{Stanford-Dog}~\cite{KhoslaYaoJayadevaprakashFeiFei_FGVC2011} contains 120 dog breeds, and the training size is 12,000 whereas the testing size is 8,580. Last, \emph{CUB200}~\cite{WahCUB_200_2011} has 200 bird species. There are 5,994 samples for training and 5,794 samples for testing.

\noindent\textbf{Setup.}
We employ two backbone models, AlexNet~\cite{alexnet} and ResNet18~\cite{He_2016}, to demonstrate the effectiveness of SUVR. 
The architectures and hyperparameters for the backbone models are identical to previous research~\cite{wu2018unsupervised} for a fair comparison.
For hyperparameters in SUVR, we set the number of neighbors being explored ($k$ in Section 2) to 4, then re-select the same neighbor size for each iteration. 
We use SGD with Nesterov~\cite{dozat2016incorporating} for the optimizer. The learning rate is set to 0.03 initially, and then is reduced 10\% for every 40 epochs. For the memory bank, the momentum of the exponential moving average was set to 0.5.

\begin{table*}[h!]
\centering
\caption{Top-1 accuracy for image classification task on five benchmark datasets.}
\centering
\begin{tabular}{ccccccccc}
\hline
Dataset          & \multicolumn{2}{c}{CIFAR-10}  & \multicolumn{2}{c}{CIFAR-100} & \multicolumn{2}{c}{SVHN}      & Stanford-Dog  & CUB-200       \\
Backbone         & AlexNet       & ResNet18      & AlexNet       & ResNet18      & AlexNet       & ResNet18      & ResNet18      & ResNet18      \\ \hline
DeepCluster       & 60.3          & 80.8          & 32.7          & 50.7          & 79.8          & 93.6          & 27.0          & 11.6          \\
ISIF             & 74.4          & 83.6          & 44.1          & 54.4          & 89.8          & 91.3          & 31.4          & 13.2          \\
SimCLR           & 73.0          & 82.3          & 43.2          & 55.8          & 88.6          & 90.8          & 33.7          & 17.5          \\
AND              & 74.8          & 84.2          & 41.5          & 56.1          & 90.9          & 94.5          & 32.3          & 14.4          \\ \hline
SUVR (BFS)    & 75.6          & 85.2          & \textbf{45.5} & \textbf{57.5} & 92.4          & 95.3          & \textbf{35.4} & \textbf{18.3} \\
SUVR (DFS)    & \textbf{76.5} & \textbf{85.4} & 44.3          & 57.3          & \textbf{92.9} & \textbf{95.8} & 34.2          & 17.8          \\
SUVR (Greedy) & 76.4          & 85.3          & 45.4          & 57.4          & 92.8          & 95.7          & 34.3          & 18.2          \\ \hline
\end{tabular}
\label{table:main}
\end{table*}

\noindent\textbf{Evaluation Plan.}
The class labels are only used to evaluate a model and were left out for training. We follow the evaluation protocol that utilizes k-nearest neighbors~\cite{knn,wu2018unsupervised} to vote for the final prediction, which is commonly used in unsupervised embedding learning task. The predicted labels are derived from a voting result: querying a test image to obtain $k$ nearest neighbors in the training set, and the prediction is the majority of its $k$ nearest neighbors' labels.

\noindent\textbf{Baselines.}
We compare our model with several strong baselines. \emph{DeepCluster}~\cite{Caron2018} is a clustering-based approach that derives pseudo-labels by the clustering \cite{1017616} results, then utilizes the pseudo-labels to train the model in a supervised-learning manner.
Both \emph{ISIF}~\cite{Ye_2019} and \emph{SimCLR}~\cite{chen2020simple} use image augmentation to form the positive pairs, and train their models in a self-supervised manner. We view \emph{AND}~\cite{huang2019unsupervised} as the main competitor because they also select negative samples from the whole image set, but the way we select negatives is different: our negative samples are selected from the positive neighbors we explored via graph traversal.
All baselines' hyperparameters are set according to the official code implementation.

\subsection{Experimental Results}

We compare the performance in terms of top-1 accuracy on the five benchmark datasets. SUVR with different search strategies are included for comparison, and the results are listed on Table~\ref{table:main}.
It is also noticeable that compared with the coarse-grained dataset, performance on the finer-grained dataset are remarkably lower for all baselines. This shows that as the similarities between classes increase, it is more challenging to have correct predictions. Here we only present the performance using ResNet18 as the backbone while AlexNet exhibits similar results.

We find that SUVR consistently outperforms all baselines by a large margin. Additionally, BFS and DFS perform better on datasets with a few classes and multiple classes, respectively.
In particular, CIFAR-100, Stanford-Dog and CUB-200 get better results on BFS, while CIFAR-10 and SVHN perform better by DFS. 
The reason that BFS outperforms DFS on multi-class datasets is that when the number of classes increases, and the granularity of labels becomes finer, it is more difficult for DFS to find the correct most similar images. In other words, the similarity between classes will increase as the number of classes increases, and this makes finding the right most similar images more difficult. 
On the other hand, when the dataset contains fewer classes, DFS shows promising results that outperforms BFS, since it is easier to find the right most similar images as the classes are distinguishable.
Moreover, the greedy search allows SUVR to achieve comparatively second-order performance among BFS and DFS. Specifically, the performance of greedy is worse than DFS but better than BFS on datasets with fewer classes, such as CIFAR-10 and SVHN, and worse than BFS but better than DFS on datasets with more classes, such as CIFAR-100, Standford-Dog and CUB-200. This is as a result of the search strategy that chooses between BFS and DFS based on the similarity gain. Such results also allow us to be free from pre-defining the search strategy but still obtain comparable performance to the greedy approach.

\subsection{Analysis of Neighbor-discovering Strategy}
\begin{figure}[!t]
    \centering
    \includegraphics[width=0.475\textwidth]{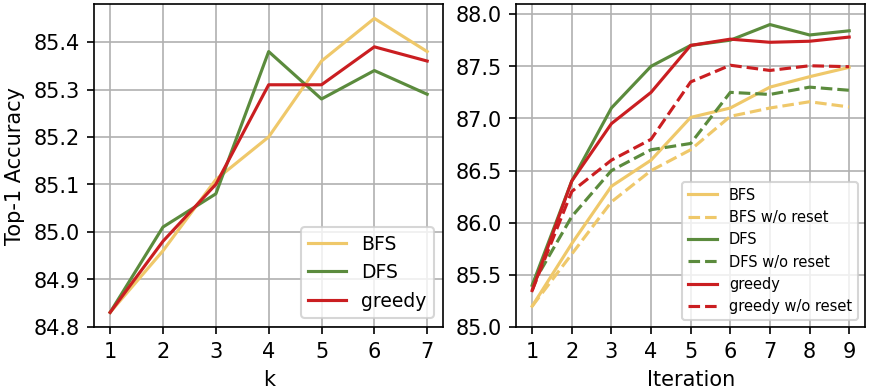}
    \caption{Left: results for varying neighbor size $k$. Right: effectiveness of resetting neighbors.}
    \label{fig:combined}
\end{figure}

\begin{figure}[!h]
    \centering
    \includegraphics[width=\linewidth]{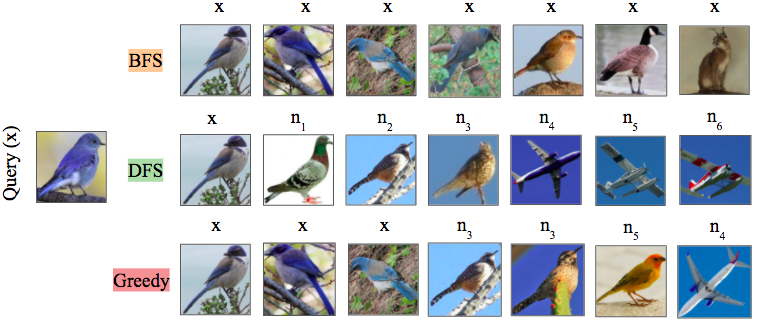}
    \caption{Illustration of neighbor discovering process.}
    \label{fig:_query}
\end{figure}

We aim to investigate the effectiveness of each neighbor-discovering strategy.
First, we analyze the effect of neighbor size $k$. In the left part of Figure~\ref{fig:combined}, we find that the accuracy increases as the neighbor size increases, especially for BFS that it can discover broader images. The accuracy of DFS drops after the neighbor size reaches 4, which might result from that DFS sequentially explores $k$-hop neighbors, and the similarity becomes lower after a certain hop number of explorations. Among BFS and DFS, the greedy search steadily remains in a second-order, showing its the robustness with a modest sacrifice on performance.
We further visualize the searching process for each strategy to discover neighbors, which is depicted in Figure~\ref{fig:_query}. Specifically, we random query an image, then plot their neighbors by searching order.
The symbol above each image is its parent image that represents the previous searching result. In particular, all searching results of BFS are based on the query image $x$, because BFS searches for top-$k$ images at one time, and the parent of DFS is the node of the previous step. The result of greedy search tends to be a tree-like structure spanning for BFS and mining for DFS. We observe that BFS finds more similar images than DFS, as DFS finds unrelated images when the neighbor size is too large (after the fourth neighbor). Such findings match with the observation found in Figure~\ref{fig:combined} that the performance using DFS drops after the neighbor size surpasses 4. The greedy search still exhibits a steady balance here, which can find more related images than DFS. 

\begin{figure}[!th]
  \centering
  \includegraphics[width=0.475\textwidth]{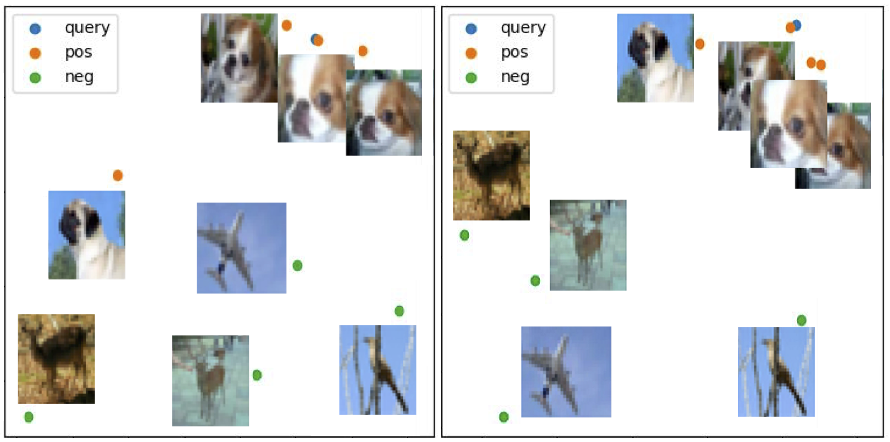}
  \caption{t-SNE visualization of \emph{SUVR} (right) and \emph{AND} (left).}
  \label{fig:tsne}
\end{figure}

\subsection{Case Study}
We conduct a case study to demonstrate the effectiveness of negative sampling. In Figure~\ref{fig:tsne}, we random sample an image, and then collect positive and negative neighbors of the image based on the greedy search.
Comparing to the competitive method \emph{AND}, the embedding of using the proposed SUVR is more visually meaningful. Concretely, the positive neighbors are more close to the query image compared with \emph{AND}, as we can see that when querying a dog image, other images of dog also appear in close positions at the top-right corner, and the negative neighbors are also separated from the query image. 
We relate this to the method we use to choose the negatives, which we select them from the positive neighbors. As a result, the negatives are somewhat similar to the image and may serve as hard negatives, which teaches SUVR how to distinguish between positive and negative samples that are very similar to the given image.
To conclude, as negative sampling is shown critical to unsupervised representation learning~\cite{chen2020simple}, our way to select negatives has shown more effective. 

\subsection{Neighbor Resetting}
Since the representation is updated every iteration, we suspect that re-selecting neighbors could consider more accurate information. On the right side of Fig~\ref{fig:combined}, we can see that re-setting neighbors is critical, as it consistently boosts the performance regardless of the search strategy.
DFS drops the most without resetting, perhaps because DFS always searches for the most similar instances, and keeping updating neighbors is especially helpful. In contrast, BFS drops the least, which could result from that BFS always searches for top-$k$ instances, lowering the impact of resetting neighbors.
\section{Conclusions}
\label{sec:conclusion}
We propose SUVR, a novel framework for unsupervised visual representation learning. We use graph traversal to explore positive neighbors and subsequently select negative samples from the found neighbors. We further present an objective function where the likelihood of positive image pairs is maximized while the likelihood of negative image pairs is minimized. Experiments conducted on five benchmark datasets demonstrate that the representations generated by SUVR are able to not only produce outstanding performance on the image classification task but also generate better representations. 

\section{Acknowledgement}
This work is supported by the National Science and Technology Council (NSTC) of Taiwan under grants 110-2221-E-006-136-MY3, 111-2221-E-006-001, 111-2634-F-002-022.

\bibliographystyle{IEEEbib}
\bibliography{paper,refs}

\begin{thebibliography}{10}

\bibitem{chen2020simple}
Ting Chen, Simon Kornblith, Mohammad Norouzi, and Geoffrey Hinton,
\newblock ``A simple framework for contrastive learning of visual
  representations,''
\newblock in {\em Proceedings of the 37th International Conference on Machine
  Learning}, 2020, ICML'20.

\bibitem{Caron2018}
Mathilde Caron, Piotr Bojanowski, Armand Joulin, and Matthijs Douze,
\newblock ``Deep clustering for unsupervised learning of visual features,''
\newblock in {\em European Conference on Computer Vision}, 2018.

\bibitem{wu2018unsupervised}
Zhirong Wu, Yuanjun Xiong, Stella~X Yu, and Dahua Lin,
\newblock ``Unsupervised feature learning via non-parametric instance
  discrimination,''
\newblock in {\em Proc. of the IEEE Conference on Computer Vision and Pattern
  Recognition}, 2018, pp. 3733--3742.

\bibitem{Ye_2019}
Mang Ye, Xu~Zhang, Pong~C. Yuen, and Shih-Fu Chang,
\newblock ``Unsupervised embedding learning via invariant and spreading
  instance feature,''
\newblock {\em 2019 IEEE/CVF Conference on Computer Vision and Pattern
  Recognition (CVPR)}, Jun 2019.

\bibitem{huang2019unsupervised}
Jiabo Huang, Qi~Dong, Shaogang Gong, and Xiatian Zhu,
\newblock ``Unsupervised deep learning by neighbourhood discovery,''
\newblock in {\em Proc. of the International Conference on Machine Learning},
  2019, pp. 2849--2858.

\bibitem{Xiao_2017}
Tong Xiao, Shuang Li, Bochao Wang, Liang Lin, and Xiaogang Wang,
\newblock ``Joint detection and identification feature learning for person
  search,''
\newblock {\em 2017 IEEE Conference on Computer Vision and Pattern Recognition
  (CVPR)}, Jul 2017.

\bibitem{li2018deeper}
Qimai Li, Zhichao Han, and Xiao-Ming Wu,
\newblock ``Deeper insights into graph convolutional networks for
  semi-supervised learning,''
\newblock in {\em Thirty-Second AAAI Conference on Artificial Intelligence},
  2018.

\bibitem{cao2016deep}
Shaosheng Cao, Wei Lu, and Qiongkai Xu,
\newblock ``Deep neural networks for learning graph representations,''
\newblock in {\em Thirtieth AAAI conference on artificial intelligence}, 2016.

\bibitem{6909475}
Ross Girshick, Jeff Donahue, Trevor Darrell, and Jitendra Malik,
\newblock ``Rich feature hierarchies for accurate object detection and semantic
  segmentation,''
\newblock in {\em 2014 IEEE Conference on Computer Vision and Pattern
  Recognition}, 2014, pp. 580--587.

\bibitem{examplebased}
Kah Sung and Tomaso Poggio,
\newblock ``Example based learning for view-based human face detection,''
\newblock {\em Pattern Analysis and Machine Intelligence, IEEE Transactions
  on}, vol. 20, pp. 39 -- 51, 02 1998.

\bibitem{felzenszwalb2010object}
Pedro~F Felzenszwalb, Ross~B Girshick, David McAllester, and Deva Ramanan,
\newblock ``Object detection with discriminatively trained part-based models,''
\newblock {\em IEEE transactions on pattern analysis and machine intelligence},
  vol. 32, no. 9, pp. 1627--1645, 2010.

\bibitem{article}
Alex Krizhevsky,
\newblock ``Learning multiple layers of features from tiny images,''
\newblock {\em University of Toronto}, 05 2012.

\bibitem{37648}
Yuval Netzer, Tao Wang, Adam Coates, Alessandro Bissacco, Bo~Wu, and Andrew~Y.
  Ng,
\newblock ``Reading digits in natural images with unsupervised feature
  learning,''
\newblock in {\em NIPS Workshop on Deep Learning and Unsupervised Feature
  Learning 2011}, 2011.

\bibitem{KhoslaYaoJayadevaprakashFeiFei_FGVC2011}
Aditya Khosla, Nityananda Jayadevaprakash, Bangpeng Yao, and Li~Fei-Fei,
\newblock ``Novel dataset for fine-grained image categorization,''
\newblock in {\em Proc. of the First Workshop on Fine-Grained Visual
  Categorization, IEEE Conference on Computer Vision and Pattern Recognition},
  Colorado Springs, CO, June 2011.

\bibitem{WahCUB_200_2011}
C.~Wah, S.~Branson, P.~Welinder, P.~Perona, and S.~Belongie,
\newblock ``{The Caltech-UCSD Birds-200-2011 Dataset},''
\newblock Tech. {R}ep. CNS-TR-2011-001, California Institute of Technology,
  2011.

\bibitem{alexnet}
Alex Krizhevsky, Ilya Sutskever, and Geoffrey Hinton,
\newblock ``Imagenet classification with deep convolutional neural networks,''
\newblock {\em Neural Information Processing Systems}, vol. 25, 01 2012.

\bibitem{He_2016}
Kaiming He, Xiangyu Zhang, Shaoqing Ren, and Jian Sun,
\newblock ``Deep residual learning for image recognition,''
\newblock {\em 2016 IEEE Conference on Computer Vision and Pattern Recognition
  (CVPR)}, Jun 2016.

\bibitem{dozat2016incorporating}
Timothy Dozat,
\newblock ``Incorporating nesterov momentum into adam,''
\newblock {\em International Conference on Learning Representations (ICLR)
  workshop}, 2016.

\bibitem{knn}
Gongde Guo, Hui Wang, David Bell, and Yaxin Bi,
\newblock ``Knn model-based approach in classification,''
\newblock {\em OTM Confederated International Conferences On the Move to
  Meaningful Internet Systems}, 08 2004.

\bibitem{1017616}
T.~{Kanungo}, D.~M. {Mount}, N.~S. {Netanyahu}, C.~D. {Piatko}, R.~{Silverman},
  and A.~Y. {Wu},
\newblock ``An efficient k-means clustering algorithm: analysis and
  implementation,''
\newblock {\em IEEE Transactions on Pattern Analysis and Machine Intelligence},
  vol. 24, no. 7, pp. 881--892, 2002.

\end{thebibliography}

\end{document}